\documentclass[runningheads]{llncs}
\usepackage{graphicx}

\usepackage{tikz}
\usepackage{comment}
\usepackage{cite}
\usepackage{amsmath,amssymb} 
\usepackage{color}

\begin{document}
\pagestyle{headings}
\mainmatter
\def\ECCVSubNumber{6697}  

\title{Learning to Generate Customized Dynamic 3D Facial Expressions}

\titlerunning{Learning to Generate Customized Dynamic 3D Facial Expressions}
\author{Rolandos Alexandros Potamias  \and
Jiali Zheng \and
Stylianos Ploumpis \and Giorgos Bouritsas \and Evangelos Ververas \and Stefanos Zafeiriou}
\authorrunning{RA Potamias et al.}
%
\institute{Department of Computing, Imperial College London, UK \\
\email{\{r.potamias,  jiali.zheng18, s.ploumpis,  g.bouritsas, e.ververas16, s.zafeiriou\}@imperial.ac.uk}\\
}
\maketitle

\begin{abstract}
 Recent advances in deep learning have significantly pushed the state-of-the-art in photorealistic video animation given a single image. In this paper, we extrapolate those advances to the 3D domain, by studying 3D image-to-video translation with a particular focus on 4D facial expressions. Although 3D facial generative models have been widely explored during the past years, 4D animation remains relatively unexplored.  To this end, in this study we employ a deep mesh encoder-decoder like architecture to synthesize realistic high resolution facial expressions by using a single neutral frame along with an expression identification. In addition, processing 3D meshes remains a non-trivial task compared to data that live on grid-like structures, such as images. Given the recent progress in mesh processing with graph convolutions, we make use of a recently introduced learnable operator which acts directly on the mesh structure by taking advantage of local vertex orderings. In order to generalize to 4D facial expressions across subjects, we trained our model using a high resolution dataset with 4D scans of six facial expressions from 180 subjects. Experimental results demonstrate that our approach preserves the subject's identity information even for unseen subjects and generates high quality expressions. To the best of our knowledge, this is the first study tackling the problem of 4D facial expression synthesis. 

\end{abstract}

\section{Introduction}

Recently, facial animation has received attention from the industrial graphics, gaming and filming communities. Face is capable to impart a wide range of information not only about the subject's emotional state but also about the tension of the moment in general. An engaged and crucial task is 3D avatar animation, which has lately become feasible \cite{ploumpis2019combining}. With modern technology, a 3D avatar can be generated by a single uncalibrated camera \cite{cao2014displaced} or even by a self portrait image \cite{ploumpis2019towards}.  At the same time, capturing facial expression is an important task in order to perceive behaviour and emotions of people. To tackle the problem of facial expression generation, it is essential to understand and model facial muscle activations that are related to various emotions. Several studies have attempted to decompose facial expressions on two dimensional spaces such as images and videos  \cite{yang2008interactive,zhou2017photorealistic,fan2019controllable,otberdout2019dynamic}. However, modeling facial expressions on high resolution 3D meshes remains unexplored. 

In contrast, few studies have attempted 3D speech-driven facial animation exclusively based on vocal audio and identity information \cite{karras_audio-driven_2017,cudeiro_capture_2019}.  Nevertheless, emotional reactions of a subject are not always expressed vocally and speech-driven facial animation approaches neglect the importance of facial expressions. For instance, sadness and happiness are two very common emotions that can  be voiced, mainly, through facial deformations. To this end, facial expressions are a major component of entertainment industry, and can convey emotional state of both scene and identity.

People signify their emotions using facial expressions in similar manners. For instance, people express their happiness by mouth and cheek deformations, that vary according to the subject’s emotional state and characteristics. Thus, one can describe expressions as “unimodal” distributions \cite{fan2019controllable}, with gradual changes from the neutral model till the apex state. Similarly to speech signals, emotion expressions are highly correlated to facial motion, but lie in two different domains.  Modeling the relation between those two domains is essential for the task of realistic facial animation. However, in order to disentangle identity information and facial expression it is essential to have a sufficient amount of data. Although most of the publicly available 3D datasets contain a large variety of facial expression, they are captured only from a few subjects.  Due to this difficulty, prior work has only focused on generating expressions in 2D.

Our aim is to generate realistic 3D facial animation given a target expression and a static neutral face. Synthesis of facial expression generation on new subjects can be achieved by expression transfer of generalized deformations \cite{thies2015real,zhou2017photorealistic}. In order to produce realistic expressions, we map and model facial animation directly on the mesh space, avoiding to focus on specific face landmarks. Specifically,  the proposed method comprises two parts: (a) a recurrent LSTM encoder to project the expected expression motion to an expression latent space, and (b) a mesh decoder to decode each latent time-sample to a mesh deformation, which is added to the neutral expression identity mesh. The mesh decoder utilizes intrinsic lightweight mesh convolutions, introduced in \cite{Bouritsas_2019_ICCV}, along with unpooling operations that act directly on the mesh space \cite{ranjan2018generating}. We train our model in an end-to-end fashion on a large scale 4D face dataset. The devised methodology tackles a novel and unexplored problem, i.e. the generation of 4D expressions given a single neutral expression mesh. Both the desired length and the target expression are fully defined and controlled by the user. Our work considerably deviates from methods in the literature as it can be used to generate 4D full-face customised expressions on real-time. Finally, our study is the first 3D facial animation framework that utilizes an intrinsic encoder-decoder architecture that operates directly on mesh space using mesh convolutions instead of fully connected layers, as opposed to \cite{cudeiro_capture_2019,karras_audio-driven_2017}.
\section{Related Work}
 
\subsubsection{Facial animation generation} Following the progress of 3DMMs, several approaches have attempted to decouple expression and identity subspaces and built linear \cite{Amberg2008,yang2011expression,bouaziz2013online} and nonlinear \cite{li_video_2017,ranjan2018generating,Bouritsas_2019_ICCV} expression morphable models. However, all of the aforementioned studies are focused on static 3D meshes and they cannot model 3D facial motion. Recently, a few studies attempted to model the relation between speech and facial deformation for the task of 3D facial motion synthesis. 
Karras et al. \cite{karras_audio-driven_2017} modeled speech formant relationships with 5K vertex positions, generating facial motion from LPC audio features. While this was the first approach to tackle facial motion directly on 3D meshes, their model is subject specific and cannot be generalized across different subjects. Towards the same direction, in \cite{cudeiro_capture_2019}, facial animation was generated using a static neutral template of the identity and a speech signal, used along with DeepSpeech \cite{hannun2014deep} to generate more robust speech features.  A different approach was utilized in \cite{pham_speech-driven_2017}, where 3D facial motion is generated by regressing on a set of action units, given MFCC audio features processed by RNN units. However, their model is trained on parameters extracted from 2D videos instead of 3D scans. Tzirakis et al. \cite{tzirakis_synthesising_2019} combined predicted blendshape coefficients with a mean face to synthesize 3D facial motion from speech, replacing also fully connected layers, utilized in previous studies, with an LSTM. Blendshape coefficients are also predicted from audio, using attentive LSTMs in \cite{tian2019audio2face}. In contrast with the aforementioned studies, the proposed method aims to model facial animations directly on 3D meshes. Furthermore, although blendshape coefficients might be easily modeled, they rely on predefined face rigs, a factor that limits their generalization to new unseen subjects.
\subsubsection{Geometric Deep Learning} Recently, the enormous amount of applications related to data residing in non-Euclidean domains motivated the need for the generalization of several popular deep learning operations, such as convolution, to graphs and manifolds. The main efforts include the reformulation of regular convolution operators in order to be applied on structures that lack consistent ordering or directions, as well as the invention of pooling techniques for graph downsampling. All relevant endeavours lie within the new research area of Geometric Deep Learning  (GDL) \cite{bronstein_geometric_2017}. The first attempts defined convolution in the spectral domain, by applying filters inspired from graph signal processing techniques \cite{shuman2013emerging}. These methods mainly boil down to either an eigendecomposition of a Graph Shift Operator (GSO) \cite{bruna}, such as the graph Laplacian, or to approximations thereof, by using polynomials \cite{defferrard2016convolutional,kipf2017semi} or rational complex functions \cite{levie2018cayleynets} of the GSO in order to obtain strict spatial localization and reduced computational complexity. Subsequent attempts generalize conventional CNNs by introducing patch operators that extract spatial relations of adjacent nodes within a local patch. To this end, several approaches generalized local patches to graph data, using geodesic  polar charts \cite{masci2015geodesic}  anisotropic diffusion operators \cite{boscaini2016learning} on manifolds or graphs \cite{atwood2016diffusion}.  MoNet \cite{monti2017geometric}, generalized previous spatial approaches by learning the patches themselves with Gaussian kernels. In the same direction, SplineCNN \cite{Fey_2018_CVPR} replaced Gaussian kernels with B-spline functions with significant speed advantage. Recent studies focused on soft-attention methods to weight adjacent nodes \cite{verma2018feastnet,velickovic2018graph}. However, in contrast to regular convolutions, the way permutation invariance is enforced in most of the aforementioned operators, inevitably renders them unaware of vertex correspondences. To tackle that, Bouritsas et al. \cite{Bouritsas_2019_ICCV} defined local node orderings, instantiated with the spiral operator of \cite{lim2018simple}, by exploiting the fixed underlying topology of certain deformable shapes, and built correspondence-aware anisotropic operators.


\subsubsection{Facial Expression datasets}
Another major reason that 4D generative models have not been widely exploited is due to the limited amount of 3D datasets. During the past decade, several 3D face databases have been published. However, most of them are static \cite{zhong2007robust,savran2008bosphorus,stratou2011effect,moreno2004gavabdb,gupta2010anthropometric}, consisted of few subjects \cite{ranjan2018generating,cosker2011facs,chang2005automatic,yin126high}, and have limited  \cite{alashkar,Fanelli2010,savran2008bosphorus} or spontaneous expressions \cite{zhang2013high,zhang2016multimodal}, making them inappropriate for tasks such as facial expression synthesis. On the other hand, the recently proposed 4DFAB dataset \cite{cheng20184dfab} consists  of six 3D dynamic facial expressions (from 180 subjects), which is ideal for subject independent facial expression generation. 
In contrast with all previously mentioned datasets, 4DFAB, due to the high range of subjects, can be a promising resource towards disentangling facial expression from the identity information. 
\section{Learnable Mesh Operators: Background}
\subsection{Spiral Convolution Networks}
We define a 3D facial surface discretized as triangular mesh $\mathcal{M} = (\mathcal{V}, \mathcal{E},\mathcal{F})$ with $\mathcal{V}$ the set of N vertices, $\mathcal{E}$ and $\mathcal{F}$ the sets of edges and faces, respectively. Let also, $X \in \mathbb{R}^{N \times d}$ denote the feature matrix of the mesh. 
In contrast to regular domains, when attempting to apply convolution operators on graph-based structures,  there does not exist a consistent way to order the input coordinates.  However, in a \emph{fixed topology} setting, such an ordering is beneficial so as to be able to keep track of the existing correspondences. In \cite{Bouritsas_2019_ICCV}, the authors identified this problem and intuitively order the vertices by using spiral trajectories \cite{lim2018simple}. In particular, given a vertex $v \in \mathcal{V}$, we can define a \emph{k-ring} and \emph{k-disk} as: 
\begin{equation}
\centering
    \begin{gathered}
    ring^{(0)}(v) = {v},\\
    ring^{(k+1)}(v) = \mathcal{N}(ring^{(k)}(v))- disk^{(k)}(v), \\
    disk^{(k)}(v) = \bigcup_{i=0,...,k} ring^{(i)}(v) 
\end{gathered}
\end{equation}
where $\mathcal{N}(\emph{S})$ is the set of all vertices adjacent to at least one vertex $\in\emph{S}$. 

Once the $ring^{(k)}$ is defined, the spiral trajectory centered around vertex \emph{v} can be defined as: 
\begin{equation}
    S(\emph{v},k) = \{ring^{(0)}(v), ring^{(1)}(v),..., ring^{(k)}(v)\}
\end{equation}
To be consistent across all vertices, one can pad or truncate S(\emph{v},k) to a fixed length $L$. To fully define the spiral ordering, we have to declare the starting direction and the orientation of a spiral sequence. In the current study, we adopt the settings followed in \cite{Bouritsas_2019_ICCV}, by selecting the initial vertex of $S(\emph{v},k)$ to be in the direction of the shortest geodesic distance between a static reference vertex. Given that all 3D faces share the same topology, spiral ordering $S(\emph{v},k)$ will be the same across all meshes and so, their calculation is done only once. With all the above mentioned, \emph{Spiral Convolution} can be defined as: 
\begin{equation}
    \mathbf{f}^*_{v} = \sum_{j=0}^{|S(v,k)|-1} \mathbf{f}(S_j(v,k)) \mathbf{W}_j 
\end{equation}
where $|S(v,k)|$ amounts to the total length of the spiral trajectory,  $\mathbf{f}(S_j(v,k))$ are the $d$-dimensional input features of the \emph{jth} vertex of the spiral trajectory, $\mathbf{f}^*$ the respective output, and $\mathbf{W}_j$ are the filter weights.

\subsection{Mesh Unpooling Operations}
In order to let our graph convolution decoder to generate faces sampled from a latent space, it is essential to use unpooling operations in analogy with transposed convolutions in regular settings. Each graph convolution is followed by an upsampling layer which acts directly on the mesh space, by increasing the number of vertices. We use sampling operations introduced in \cite{ranjan2018generating}, based on sparse matrix multiplications with upsampling matrices $Q_u \in \{0,1\}^{n\times m}$, where $m>n$. Since upsampling operation changes the topology of the mesh, and in order to retain the face structure, upsampling matrices $Q_u$ are defined on the basis of down-sampling matrices. The barycentric coordinates of the vertices that were discarded during downsampling procedure are stored and used as the new vertex coordinates of the upsampling matrices. 

\section{Model}
\begin{figure}
    \centering
  \includegraphics[scale=0.4]{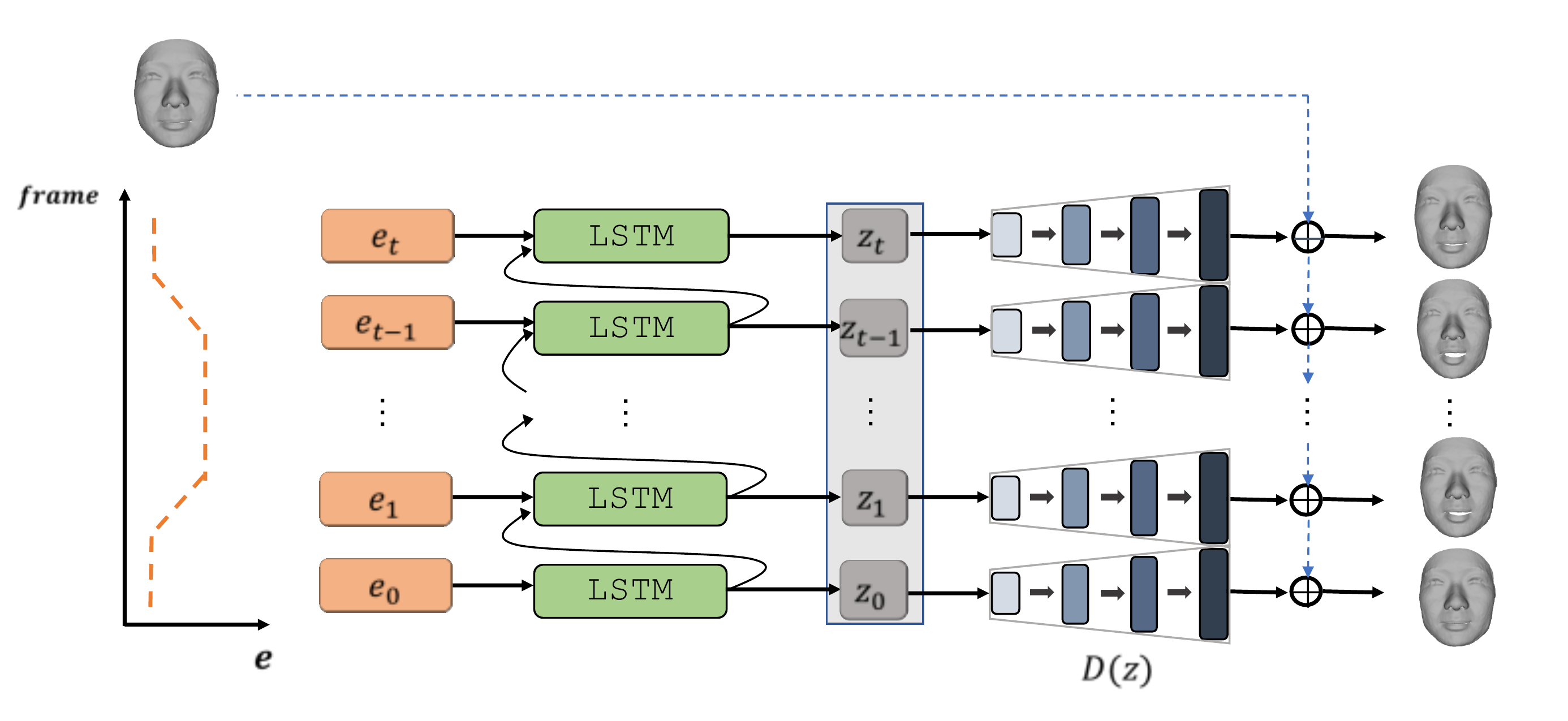}
    \caption{Network architecture of the proposed method.}
    \label{model}
\end{figure}

The overall architecture of our model is structured by two major components (see Figure \ref{model}). The first one contains a temporal encoder, using an LSTM layer that encodes the expected facial motion of the target expression. It takes as input a temporal signal $e \in R^{6 \times T}$ with length $T$, equal to the target facial expression, also equipped with information about the time-stamps that show when the generated facial expression should reach onset, apex and offset modes. Each time-frame of signal $e$ can be characterised as a one-hot encoding of one of the six expressions, with amplitude that indicates the scale of the expression.  The second component of our network consists of a frame decoder, with four layers of mesh convolutions, where each one is followed by an upsampling layer. Each upsampling layer increases the number of vertices by five times, and every mesh convolution is followed by a ReLU activation \cite{nair2010rectified}. Finally, the output of the decoder is added to the identity neutral face. Given a time sample from the latent space the frame decoder network models the expected deformations on the neutral face. Each output time frame can be expressed as: 

\begin{equation}
\centering
\begin{gathered}
    \hat{x}_t = D(z_t) + x_{id}, \\
    z_t = E(e_t)
\end{gathered}
\end{equation}
where $D(\cdot)$ denotes the mesh decoder network, $E(\cdot)$ the LSTM encoder, $e_t$ the facial motion information for time-frame $t$ and $x_{id}$ the neutral face of the identity. 
The network details can be found in Table \ref{decoderParam}. We trained our model for 100 epochs with learning rate of 0.001 and a weight decay of 0.99 on every epoch. We used Adam optimizer \cite{kingma2014adam} with a 5e-5 weight decay. 

\subsubsection{Loss function.} The mesh decoder network outputs motion deformation for each time-frame with respect to the expected facial animation. To train our model we minimize both the reconstruction error $L_r$ and the temporal coherence $L_c$, as proposed in \cite{karras_audio-driven_2017}.  Specifically, we define our loss function between the generated time frame $\hat{x}_t$ and its ground truth $x_t$ value as: 
\begin{equation}
\label{loss}
\centering
\begin{gathered}
    L_r(\hat{x}_t,x_t) = \left\lVert \hat{x}_t-x_t\right\rVert_{1} \\
    L_c(\hat{x}_t,x_t) = \left\lVert(\hat{x}_t-\hat{x}_{t-1})- (x_t-x_{t-1})\right\rVert_{1} \\
    L(\hat{x}_t,x_t) =  L_r(\hat{x}_t,x_t) + L_c(\hat{x}_t,x_t)
\end{gathered}
\end{equation}
Although reconstruction loss $L_r$ term can be sufficient to encourage model to match ground truth vertices at each time step, it does not produce high-quality realistic animation. On the contrary, temporal coherence loss $L_c$ term ensures temporal stability of the generated frames by matching the distances between consecutive frames on ground truth and generated expressions.
\begin{table}[!h]
\centering
\caption{Mesh Decoder architecture}
\begin{tabular}{lcc}
Layer                               & Input Dimension & \multicolumn{1}{l}{Output Dimension} \\ \hline \hline
\multicolumn{1}{c}{Fully Connected} & 64                       & 46x64                                         \\
Upsampling                          & 46x64                    & 228x64                                        \\
Convolution                         & 228x64                   & 228x32                                        \\
Upsampling                          & 228x32                   & 1138x32                                       \\
Convolution                         & 1138x32                  & 1138x16                                       \\
Upsampling                          & 1138x16                  & 5687x16                                       \\
Convolution                         & 5687x16                  & 5687x8                                        \\
Upsampling     & 5687x8                   & 28431x8                                       \\
Convolution   & 28431x8                  & 28431x3 \\                         \hline            
\end{tabular}
\label{decoderParam}
\end{table}
\section{Experiments}
\subsection{Dynamic 3D face database}
\label{dataset}
To train our expression generative model we use the recently published 4DFAB \cite{cheng20184dfab}. 4DFAB contains dynamic 3D meshes of 180 people (60 females, 120 males) with ages between 5 to 75 years. The devised meshes display a variety of complex and exaggerated facial expressions, namely \textit{happy, sad, surprise, angry, disgust} and \textit{fear}. The 4DFAB database displays high variance in terms of ethnicity origins, including subjects from more than 30 different ethnic groups. We split the dataset into 153 subjects for training and 27 for testing. The data were captured with 60fps, thus each expression is sampled every approximately 5 frames in order to allow our model to generate extreme facial deformations. Given the high quality of the data (each mesh is composed by 28K vertices) as well as the relatively big number of subjects, 4DFAB presents a rich and rather challenging choice for training generative models. 

\subsubsection{Expression Motion Labels} 
\begin{figure}[!ht]
    \centering
        \includegraphics[scale=0.5]{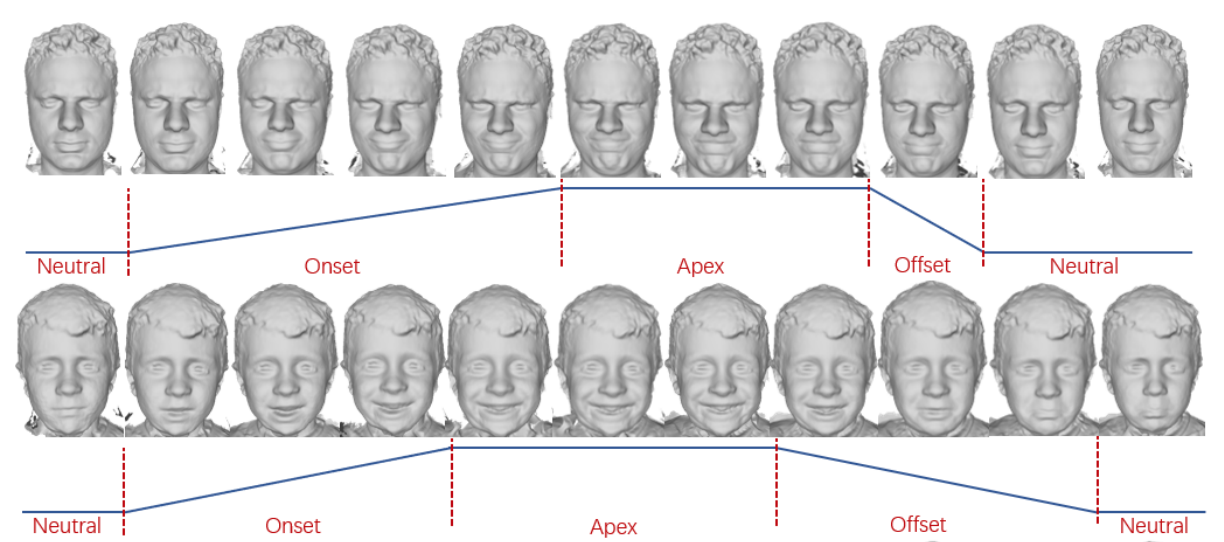}
    
    \caption{Sample subjects from the 4DFAB database posing an expression along with expression motion labels.}
   \label{fig:annotation}
\end{figure}

In this study, we rely on the assumption that each expression can be characterised by four phases of its evolution (see Figure \ref{fig:annotation}). First, the subject starts from a neutral pose and at a certain point their face starts to deform, in order to express their emotional state. We call this the \textit{onset phase}. After the subject’s expression reaches its \textit{apex state}, it will start again its deformation from the peak emotional state until it becomes neutral again. We call this the \textit{offset phase}. Thus, each time frame is assigned a label that reflects its emotional state phase. We consider the emotional state as a value ranging from 0 to 1 assigned to each frame, with 0 representing the neutral phase and 1 the apex phase. Onset and offset phases are represented via a linear interpolation between the apex and neutral phases (see Figure \ref{fig:annotation}). 
However, expressions may also range in terms of extremeness, i.e. the level of intensity in subject's expression. To let our model learn diverse extremeness levels for each expression, it is essential to scale each expression motion label from $[0,1]$ to $[0,s_i]$, where $s_i \in (0,1]$ represents the scaled value of the apex state according to the intensity of the expression. Intuitively, the extremeness of each expression is proportional to the absolute mean deformation of the expression, we can thus calculate scaling factor $s_i$ as:
\begin{equation}
    \centering
     s_i = \frac{clip(\frac{m_i - \mu_e}{\sigma_e})+1}{2}
\end{equation}
where $m_i$ is a scalar value representing the absolute value of the mean deformation of the sequence from neutral frame and $\mu_e$, $\sigma_e$ the mean and standart deviation of the deformation of the respective expression. Clip() function is used to clip values to [-1,1].

\subsection{Dynamic Facial Expressions}
	The proposed model for the generation of facial expressions is assessed both qualitatively and quantitatively. The model is trained by feeding the neutral frame of each subject and the manifested motion (i.e. the time-frames where the expression reaches onset, apex and offset modes) of the target expression.
	We evaluated the performance of the proposed model by its ability to generate expressions of 27 unobserved test subjects. To this end, we calculated the reconstruction loss as the per-vertex Euclidean distance between each generated sample and its corresponding ground truth. 
	
	\textbf{Baseline.} As a comparison baseline we implemented an expression blendshape decoder that transforms the latent representation $z_t$ of each time-frame, i.e. the LSTM outputs, to an output mesh. In particular, expression blendshapes were modeled by first subtracting the neutral face of each subject to its corresponding expression,  for all the corresponding video frames. With this operation we are able to capture and model just the motion deformation of each expression. Then, we applied Principal Component Analysis (PCA) to the motion deformations to reduce each expression to a latent vector. For a fair comparison, we use the same latent size for both the baseline and the proposed method. 
	
The results presented in Table \ref{generation_loss} show that the proposed model outperforms the baseline with regards to all the expressions, as well as on the entire dataset (0.39mm vs 0.44mm). 
\begin{table}[!ht]
\centering
\caption{Generalization per-vertex loss over all expressions, along with the total loss.}
\label{generation_loss}
\begin{tabular}{l|cccccc|l}
Model    & \textit{Happy} & \textit{Angry} & \textit{Sad} & \textit{Surprise} & \textit{Fear} & \textit{Disgust} & Total \\ \hline \hline
Baseline &     0.49 & 0.37   & 0.37        &      0.48 &        0.43  & 0.45        &    0.44  \\
\textbf{Proposed} &    \textbf{0.37}  &  \textbf{0.35}  & \textbf{0.36} &  \textbf{0.43} &  \textbf{0.42}&  \textbf{0.42}&   \textbf{0.39}    \\ \hline
\end{tabular}
\end{table}

\begin{figure}[!h]
    \centering 
        \includegraphics[scale=0.22]{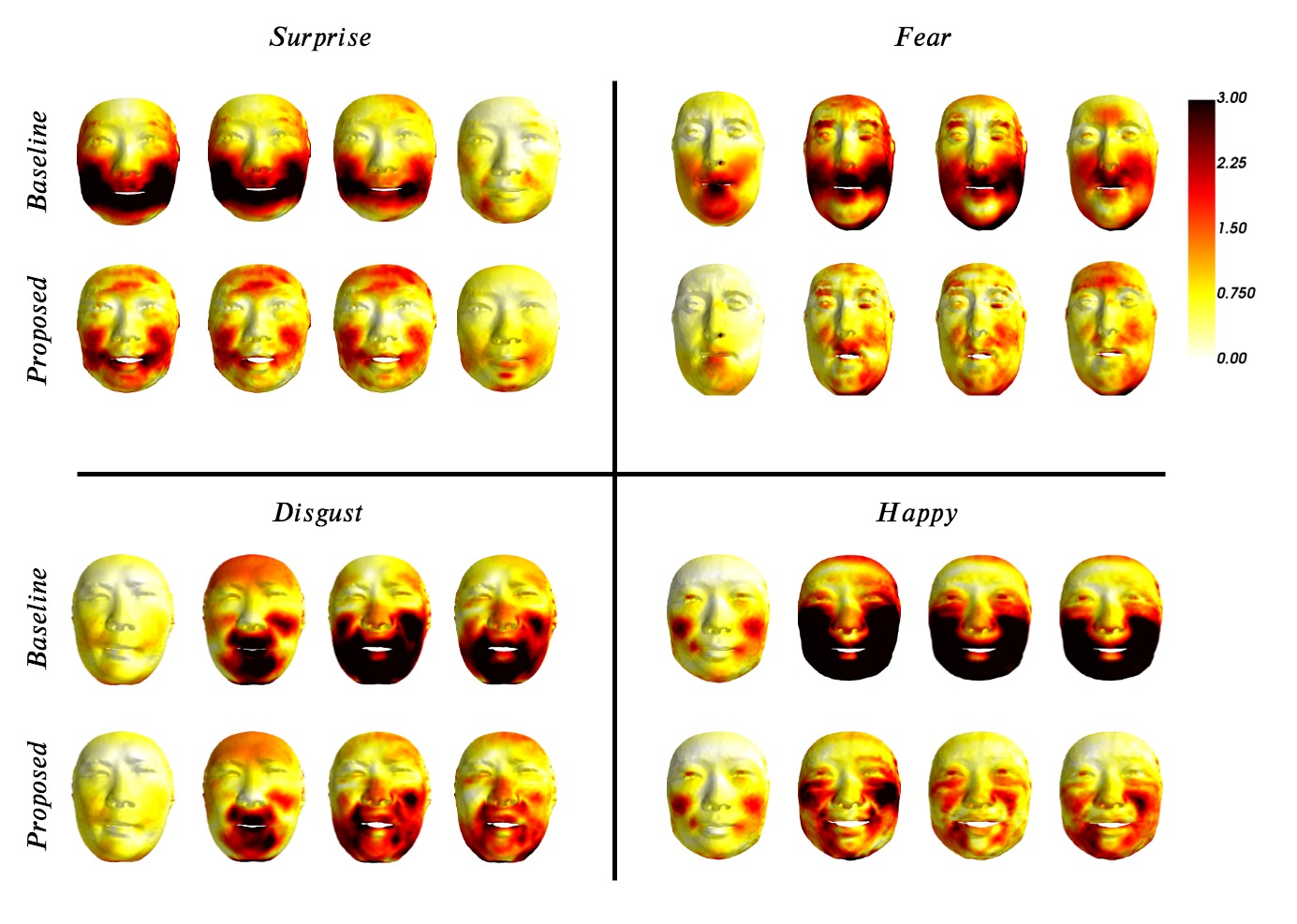}
    \caption{Color heatmap visualization of error metric of both baseline (top rows) and proposed (bottom rows) model against the ground truth test data for four different expressions.}
   \label{fig:heatmap}
\end{figure}

Moreover, as can be seen in Figure \ref{fig:heatmap}, the proposed method can produce more realistic animation, especially in the mouth region, compared to PCA blendshapes. Error visualizations in Figure \ref{fig:heatmap}, show that the blendshape model produces mild transitions between each frame and cannot generate extreme deformations. Note also that the proposed method errors are mostly centered around the mouth and the eyebrows, due to the fact that our model is subject independent and  each identity expresses its emotions in different ways and varying extents. In other words, the proposed method models each expression with respect to the motion labels without taking into account identity information, thus the generated expressions can have some variations compared to the ground truth subject-dependent expression. 

\begin{figure}[!b]
    \centering
        \includegraphics[scale=0.14]{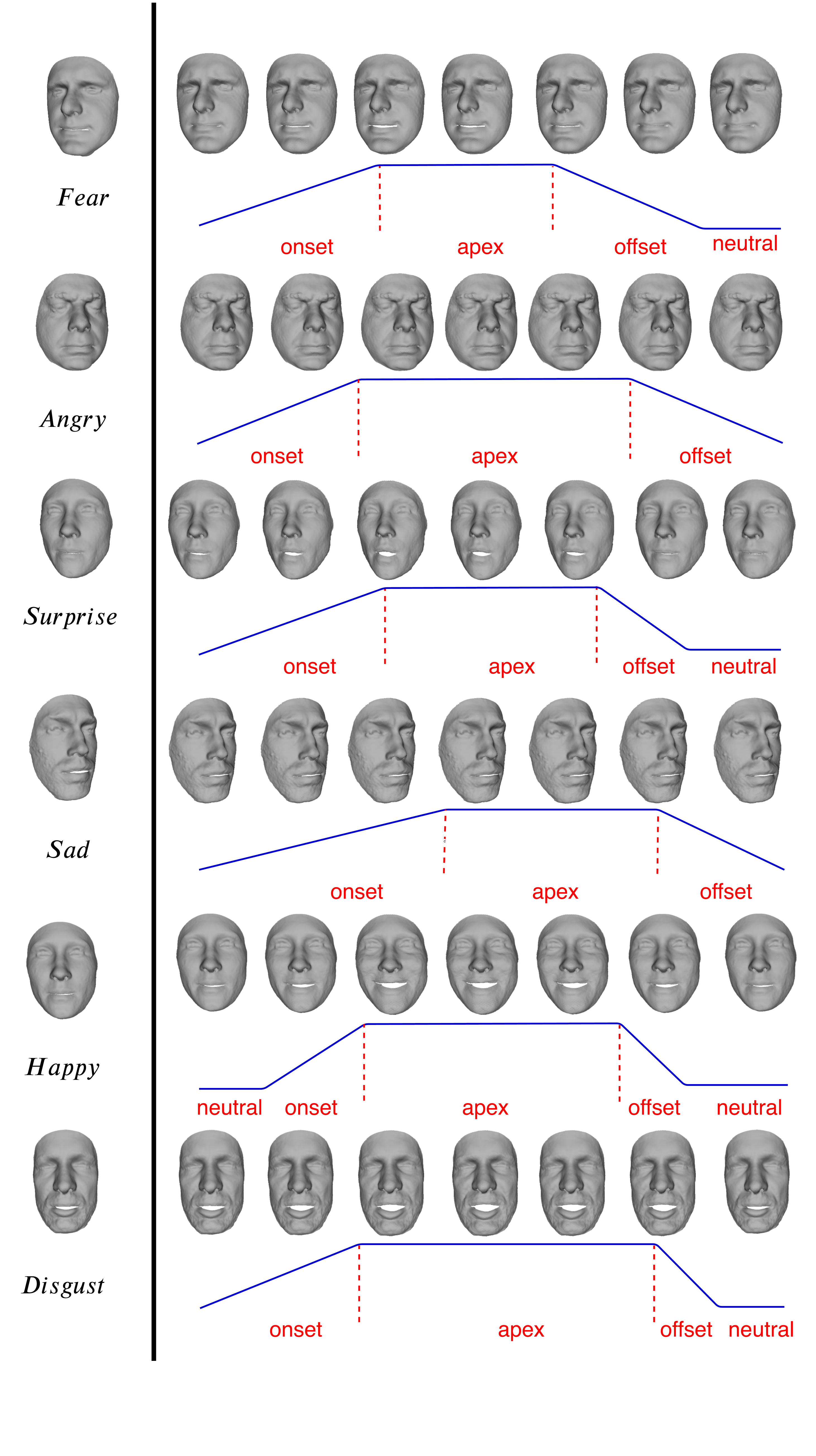}
    
    \caption{Frames of generated expressions along with their expected motion labels: Fear, Angry, Surprise, Sad, Happy, Disgust (from top to bottom).}
   \label{fig:generated_images}
\end{figure}

For a subjective qualitative assessment, Figure \ref{fig:generated_images} shows several expressions that were generated by the proposed method. Due to the fact that several expressions, such as angry and sad expression, mainly relate to eyebrow deformations can not be easily visualised by still images we encourage the reader to check our supplementary material for more qualitative results.

\subsection{Classification of generated 4D expressions}
To further assess the quality of the generated expressions we implemented a classifier trained to identify expressions. 
The architecture of the classifier is based on a projection on a PCA space followed by three fully connected layers. The sequence classification is performed in two steps. In particular, first we computed 64-PCA coefficients to represent all expression and deformation variations of the training set and then we used them as a frame encoder to map the unseen test data to a latent space. Following that, the latent representations of each frame are concatenated and processed by fully-connected layers in order to predict the expression of the given sequence. The network was trained on the same training set that was originally used for the generation of expressions, with Adam optimizer and 5e-3 weight decay for 13 epochs. Table \ref{classification} shows the achieved classification performance for the ground truth test data and the generated data from the proposed model and from the baseline model, respectively. 

\begin{table}[!ht]
\centering
\caption{Constructing classification performance between ground truth (test) data, generated (by the proposed method) data, and the blendshape baseline. }
\label{classification}
\begin{tabular}{l|lll|lll|lll|}
         & \multicolumn{3}{c|}{Baseline}                                                                          & \multicolumn{3}{c|}{\textbf{Proposed}}                                                                         & \multicolumn{3}{c|}{Ground Truth}                                                                     \\
         &  \multicolumn{1}{c}{Pre}  & \multicolumn{1}{c}{Rec} & \multicolumn{1}{c|}{F1} &  \multicolumn{1}{c}{Pre} & \multicolumn{1}{c}{Rec} & \multicolumn{1}{c|}{F1}  & \multicolumn{1}{c}{Pre} & \multicolumn{1}{c}{Rec} & \multicolumn{1}{c|}{F1} \\ \hline \hline
Surprise & 0.69  & 0.62 & 0.65  			& \textbf{0.74}  & \textbf{0.90} & \textbf{0.81}            & 0.69  & 0.83 & 0.75  \\
Angry    & 0.67  &0.55  & 0.60 	& 0.75   & 0.52  & \textbf{0.61}                                                    & \textbf{0.81} & \textbf{0.59}  & 0.60  \\
Disgust  & 0.51 & 0.72  & 0.60  			& \textbf{0.65}  & \textbf{0.83} & \textbf{0.73}            & 0.58 & 0.72  & 0.65   \\
Fear     & 0.59  & 0.36  & 0.44   			& \textbf{0.67}  & \textbf{0.43}& \textbf{0.52}             & 0.64  & 0.32 & 0.43    \\
Happy    & 0.63 & 0.59   & 0.60 			& 0.71              & 0.86         & 0.78                            & \textbf{0.73} &\textbf{0.93} & \textbf{0.82}   \\
Sad      & 0.43  & 0.57 & 0.49   			& 0.62              & 0.60          &0.61                            & \textbf{0.74}  & \textbf{0.77} & \textbf{0.75} \\ \hline \hline
Total  &  0.58   & 0.57 & 0.57  			& 0.69  & 0.69 & 0.68                                                 & \textbf{0.70}  & \textbf{0.70}  & \textbf{0.68}                    \\ \hline
\end{tabular}
\end{table}
As can be seen in Table \ref{classification}, the generated data from the proposed model achieve similar classification performance with ground truth data across almost every expression. In particular, the generated \textit{surprise, disgust} and \textit{fear} expressions from the proposed model can be even easier classified compared to the ground truth test data. Note also that both ground truth data and the data generated by the proposed model achieve 0.68 F1-score. 

\subsection{Loss per frame}
\begin{figure}[!ht]
    \centering
  \includegraphics[scale=0.5]{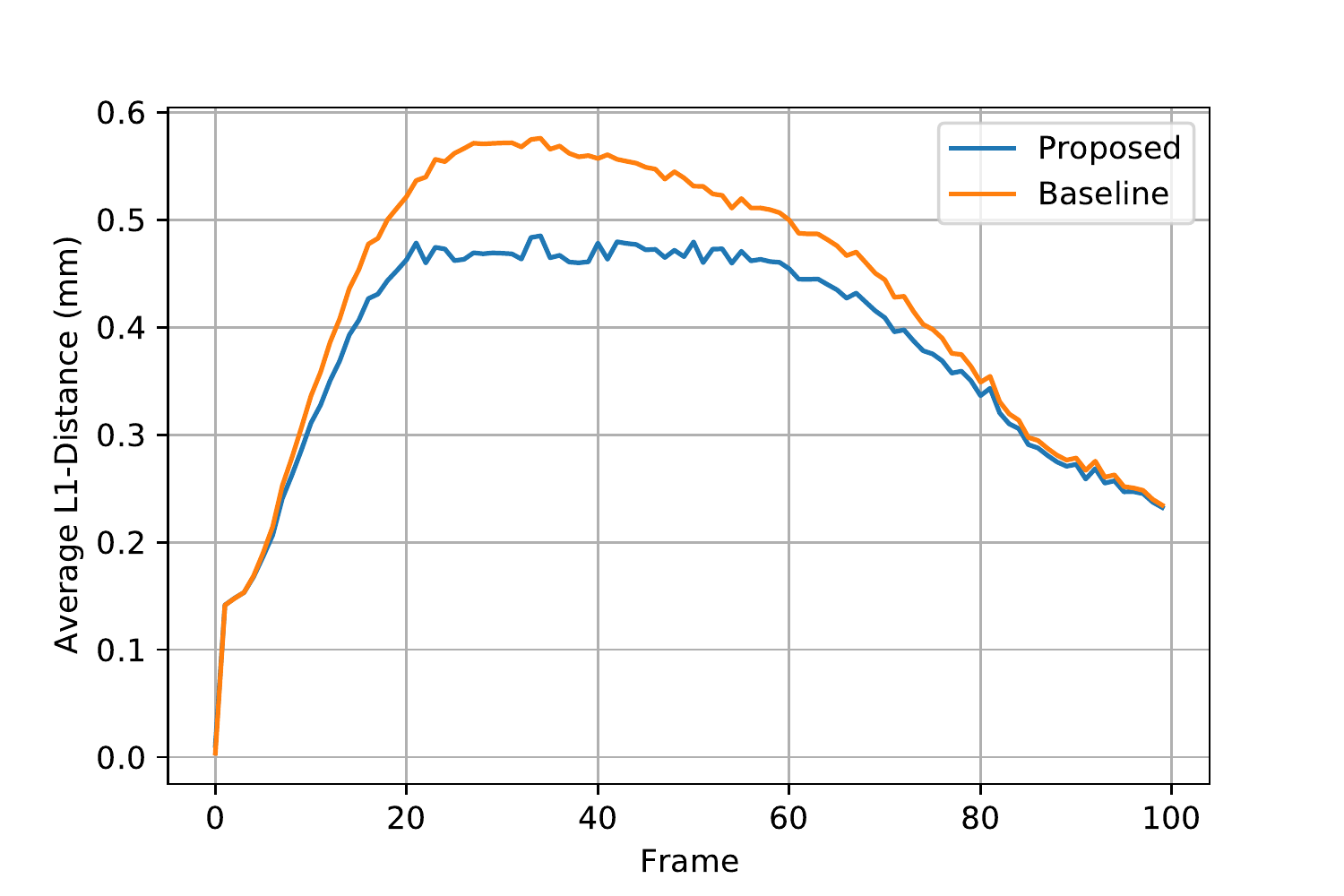}
    \caption{Average per-frame $L_1$ error between the proposed method and the PCA-based blendshape baseline.}
    \label{fig:frame_loss}
\end{figure}

Since the overall loss is calculated for all frames of the generated expression, we cannot assess the ability of the proposed model to generate with low error-rate the onset, apex and offset of each expression. To evaluate the performance of the model on each expression phase we calculated the average $L_1$ distance between the generated and the corresponding ground truth frame for each of the frames of the evolved expression. Figure \ref{fig:frame_loss} shows that apex phase, which is usually taking place between time-frames 30-80, has an increased $L_1$ error for both models. However, the proposed method exhibits a stable loss around 0.45mm across the apex phase, compared to the blendshape baseline that struggles to model the extreme deformations that take place and characterize the apex phase of the expression.

\subsection{Interpolation on the latent space}
\begin{figure}[!ht]
    \centering
  \includegraphics[scale=0.11]{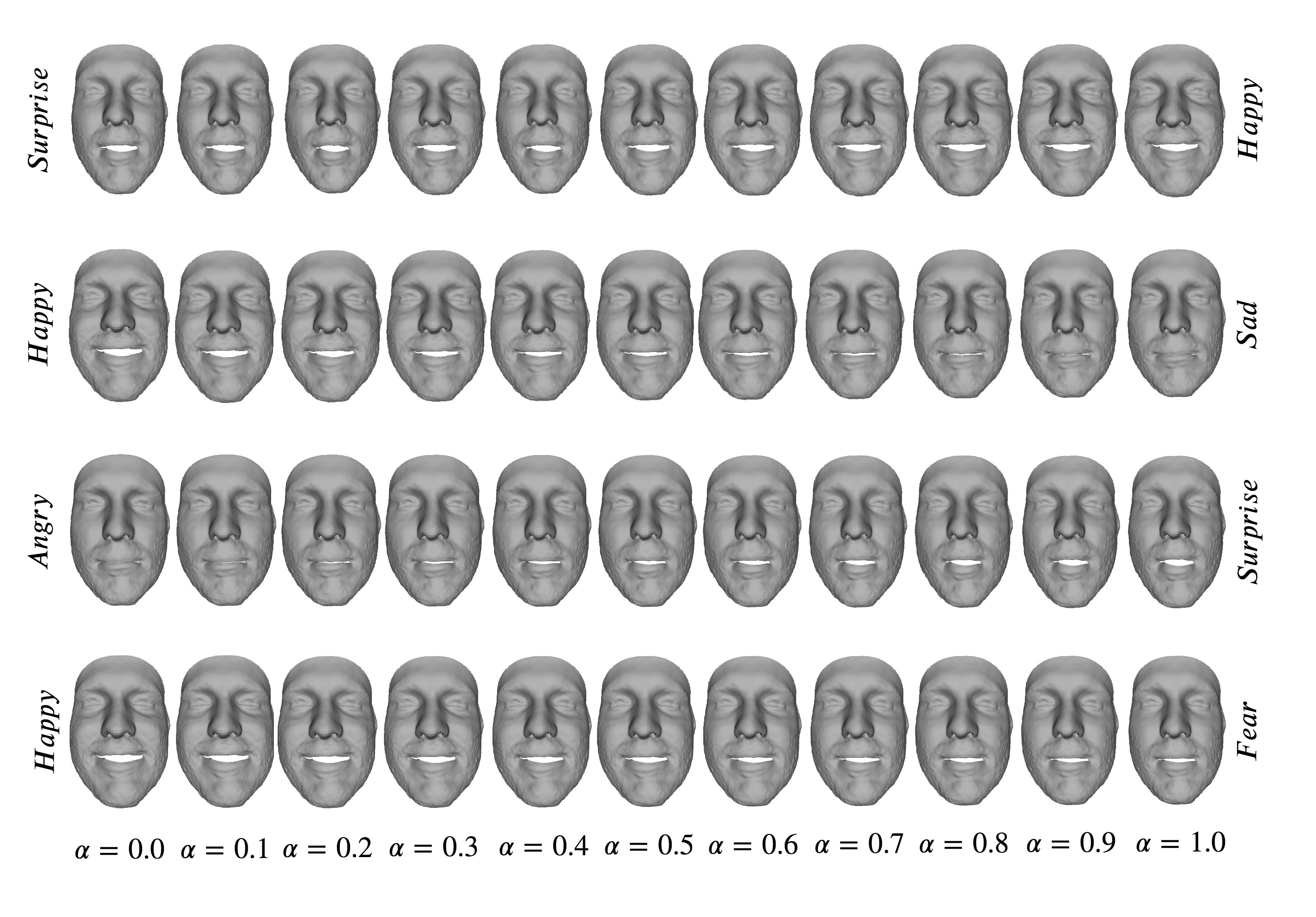}
    \caption{Interpolation on the latent space between different expressions.}
    \label{fig:interpolation}
\end{figure}

To qualitatively evaluate the representation power of the proposed LSTM encoder we applied linear interpolation to the expression latent space. Specifically, we choose two different apex expression labels from our test set, we encode them using our LSTM encoder to two latent variables $z_0$ and $z_1$, each one of size 64.  We then produce all intermediate encodings by linearly interpolating the line between them, i.e. $z_{\alpha} = \alpha z_1 + (1-\alpha)z_0$, $\alpha \in (0,1)$. The latent samples $z_{\alpha}$ are then fed to our mesh decoder network. We visualize the interpolations between different expressions in Figure \ref{fig:interpolation}. 

\subsection{Expression generation in-the-wild}
\begin{figure}[!h]
    \centering
  \includegraphics[scale=0.08]{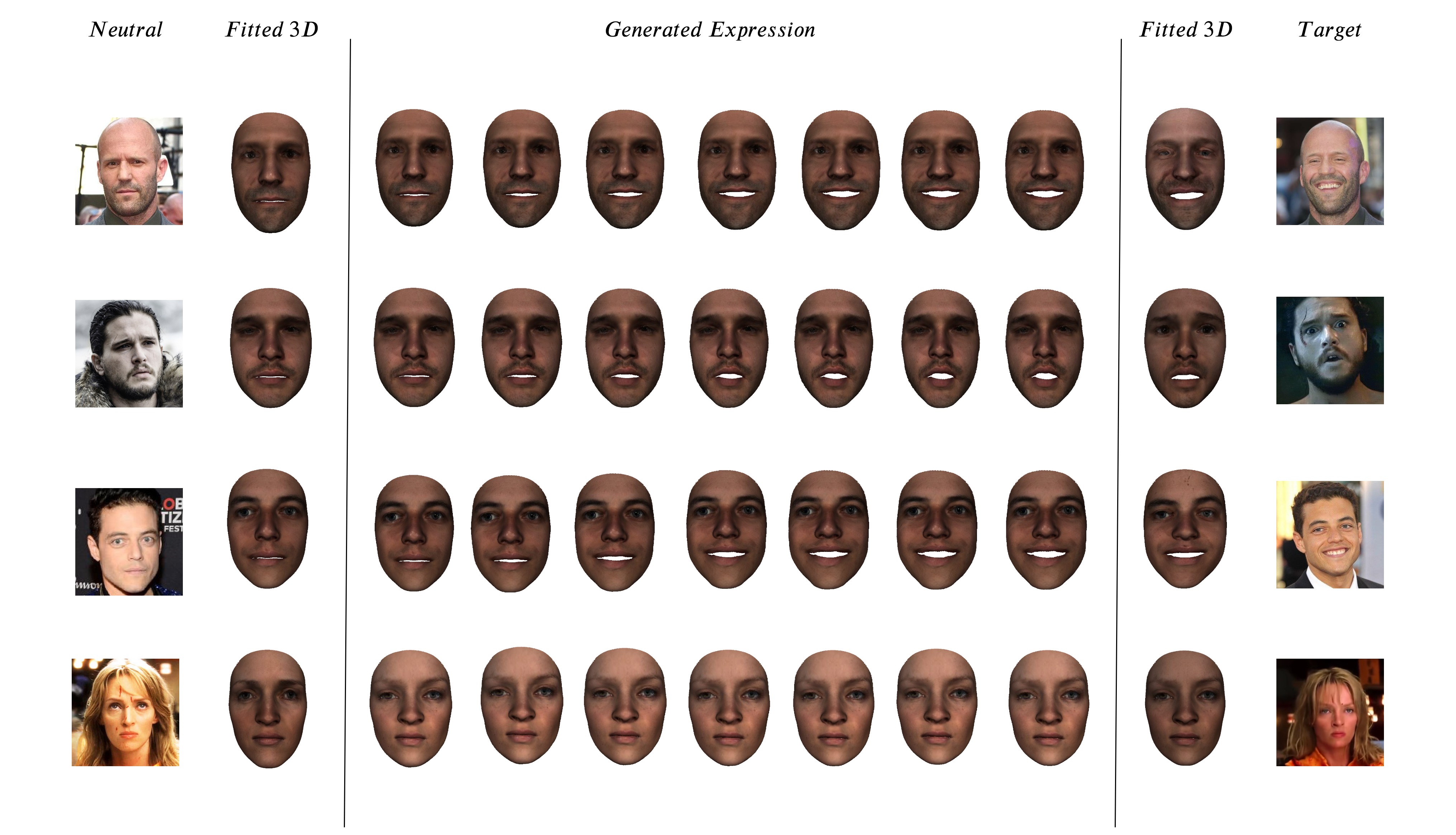}
    \caption{Generation of expressions in-the-wild form 2D images.}
    \label{fig:actors}
\end{figure}

Since expression generation is an essential task in graphics and film industries, we propose a real-world application of our 3D facial expression generator. In particular, we have collected several image pairs with neutral and various expressions of the same identity and we have attempted to realistically synthesize the 4D animation of the target expression solely relying on our proposed approach. In order to acquire a neutral 3D template for our animation purposes, we applied a fitting methodology in the neutral image as proposed in \cite{gecer2019ganfit}. By utilizing the fitted neutral mesh, we applied our proposed model in order to generate several target expressions as shown in Figure \ref{fig:actors}. Our method is able to synthesize/generate a series of realistic 3D facial expressions which demonstrate the ability of our framework to animate a template mesh given a desired expression. We can qualitatively evaluate the similarity of the generated expression with the target expression of the same identity by comparing the generated mesh with the fitted 3D face.

\section{Limitations and Future Work}
Although the proposed framework is able to model and generate a wide range of realistic expressions, it cannot model thoroughly extremeness variations. As mentioned in section \ref{dataset}, we attempted to adapt the extremeness variations of each subject into the training procedure by using an intuitive scaling trick. However, it's not certain that the mean absolute deformation of each mesh always represents the extremeness of the conducted expression. 
In addition, in order to synthesize realistic facial animation, it is essential to model along with shape deformations also facial wrinkles of each expression.  Thus, we will attempt to generalize facial expression animation to both shape and texture, by extrapolating our model to texture prediction.

\section{Conclusion}
In this paper, we propose the first generative model to synthesize 3D dynamic facial expressions from a still neutral 3D mesh. Our model captures both local and global expression deformations using graph based upsampling and convolution operators. Given a neutral expression mesh of a subject and a time signal that conditions the expected expression motion, the proposed model generates dynamic facial expressions for the same subject that respects the time conditions and the anticipated expression. The proposed method models the animation of each expression and deforms the neutral face of each subject according to a desired motion. Both expression and motion can be fully defined by the user. Results show that the proposed method outperforms expression blendshapes and creates motion-consistent deformations, validated both qualitatively and quantitatively. In addition, we assessed whether the generated expressions can be correctly classified and identified by a classifier trained on the same dataset. Classification results endorse our qualitative results showing that the generated data can be similarly classified, compared to the ones created by blendshapes model. In summary, the proposed model is the first that attempts and manages to synthesize realistic and high-quality facial expressions from a single neutral face input.

\section{Acknowledgements}
S. Zafeiriou and J. Zheng acknowledge funding from the EPSRC Fellowship DEFORM: Large Scale Shape Analysis of Deformable Models of Humans (EP/S010203/1). R.A. Potamias and G. Bouritsas were funded by the Imperial College London, Department of Computing, PhD scholarship.

%
%
\bibliographystyle{splncs04}
\bibliography{egbib}
\end{document}